\documentclass[10pt,twocolumn,letterpaper]{article}

\usepackage{cvpr}              

\usepackage{graphicx}
\usepackage{amsmath}
\usepackage{amssymb}
\usepackage{booktabs}
\usepackage{CJKutf8}
\usepackage{multicol}
\usepackage{tabularx, makecell, multirow} 

\usepackage[pagebackref,breaklinks,colorlinks]{hyperref}

\usepackage[capitalize]{cleveref}

\begin{document}

\title{Nonuniform-to-Uniform Quantization: Towards Accurate Quantization via Generalized Straight-Through Estimation}
\author{Zechun Liu$^{1,2,4}$ \hspace{0.3cm} Kwang-Ting Cheng$^1$ \hspace{0.3cm} Dong Huang$^2$ \hspace{0.3cm} Eric Xing$^{2,3}$ \hspace{0.3cm} Zhiqiang Shen$^{2,3}$\\
$^1$Hong Kong University of Science and Technology \hspace{0.3cm} $^2$Carnegie Mellon University \\ 
$^3$Mohamed bin Zayed University of Artificial Intelligence \hspace{0.3cm} 
$^4$Reality Labs, Meta Inc.\\
{\tt \small \{zliubq, timcheng\}@ust.hk \hspace{0.2cm} epxing@cs.cmu.edu \hspace{0.2cm} \{dghuang,zhiqians\}@andrew.cmu.edu }}
\maketitle

\begin{abstract}
The nonuniform quantization strategy for compressing neural networks usually achieves better performance than its counterpart, \textit{i.e.}, uniform strategy, due to its superior representational capacity. However, many nonuniform quantization methods overlook the complicated projection process in implementing the nonuniformly quantized weights/activations, which incurs non-negligible time and space overhead in hardware deployment. In this study, we propose Nonuniform-to-Uniform Quantization (N2UQ), a method that can maintain the strong representation ability of nonuniform methods while being hardware-friendly and efficient as the uniform quantization for model inference. We achieve this through learning the flexible in-equidistant input thresholds to better fit the underlying distribution while quantizing these real-valued inputs into equidistant output levels. To train the quantized network with learnable input thresholds, we introduce a generalized straight-through estimator (G-STE) for intractable backward derivative calculation \textit{w.r.t.} threshold parameters. Additionally, we consider entropy preserving regularization to further reduce information loss in weight quantization. Even under this adverse constraint of imposing uniformly quantized weights and activations, our N2UQ outperforms state-of-the-art nonuniform quantization methods by $0.5\!\!\sim\!\!1.7$\% on ImageNet, demonstrating the contribution of N2UQ design. Code and models are available at: \url{https://github.com/liuzechun/Nonuniform-to-Uniform-Quantization}.

\end{abstract}

\section{Introduction}
\label{sec:intro}

Deep Neural Networks (DNNs) have demonstrated great success in various real-world applications~\cite{he2016deep,simonyan2014very}. Despite their remarkable results, the large model size and high computational cost hinder pervasive deployment of DNNs, especially on resource-constrained devices. A number of approaches have been proposed to compress and accelerate DNNs, including channel pruning~\cite{liu2019metapruning,liu2017learning}, quantization~\cite{zhang2018lq,zhou2016dorefa,liu2018bi}, neural architecture search~\cite{cai2018proxylessnas,wu2019fbnet}, etc. 

\begin{figure}[t]
\includegraphics[width=\linewidth]{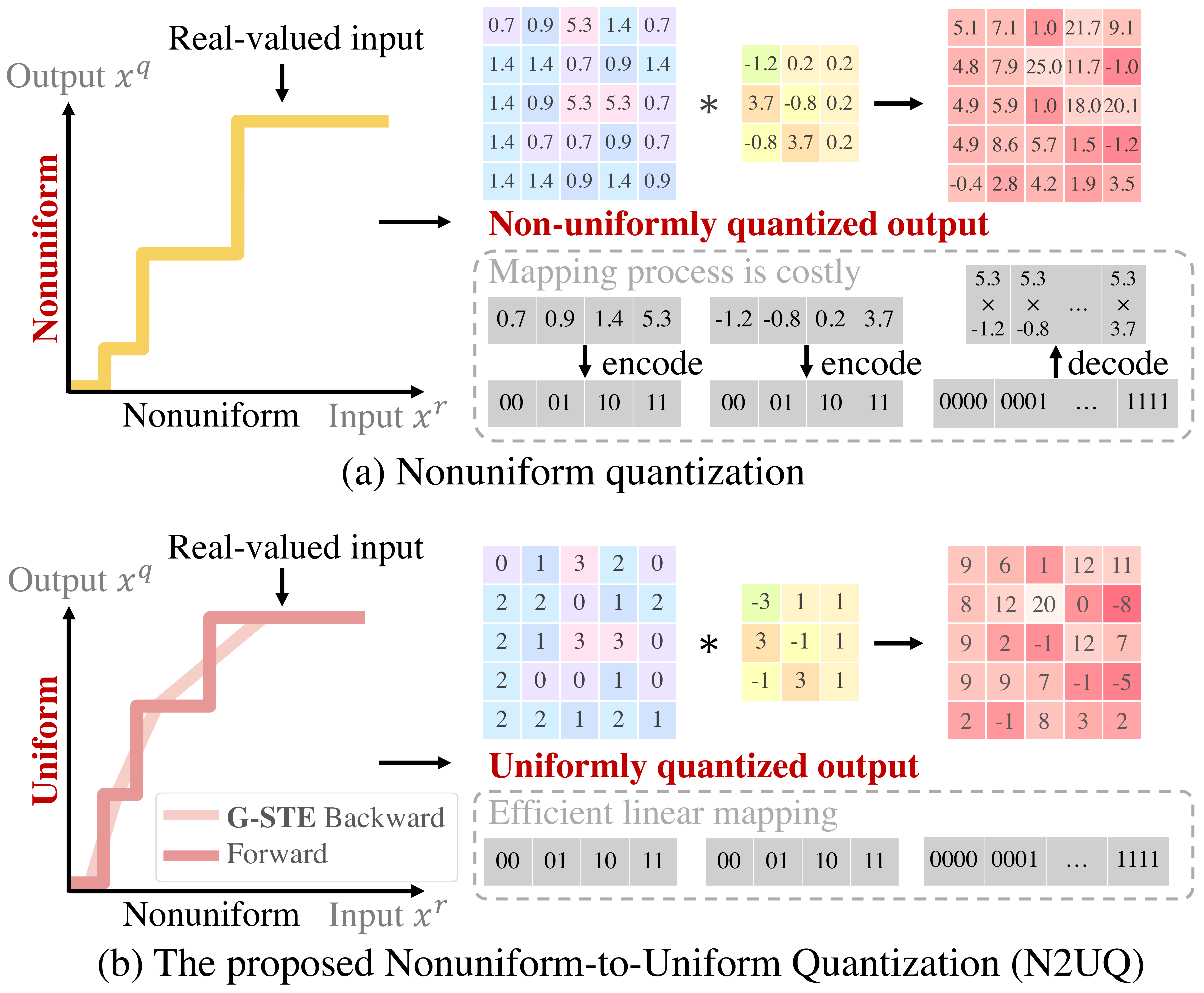}
\vspace{-1em}
\caption{(a) Previous nonuniform quantization function outputs weights and activations in in-equidistant levels, which requires the post-processing of mapping floating-point levels to binary digits in order to obtain the speed-up effect of quantization~\cite{zhou2016dorefa,gholami2021survey,arish2015efficient}.
(b) The proposed N2UQ learns input thresholds to allow more flexibility, while outputs uniformly quantized values, enabling hardware-friendly linear mappings and efficient bitwise operations. The intractable gradient computation \textit{w.r.t.} input thresholds is tackled with the proposed generalized straight-through estimator (G-STE).}
\label{fig:overall}
\end{figure}

Among these methods, quantization-based methods have shown promising results in compressing the model size by representing weights with fewer bits, and faster inference by replacing computationally-heavy convolution operations with efficient bitwise operations~\cite{rastegari2016xnor,zhang2018lq}.
Despite these advantages, quantized DNNs still have a non-negligible performance gap from their full-precision counterparts, especially with extremely low-bit quantization. For example, the 2-bit classic uniformly quantized ResNet-50 achieves 67.1\% top-1 accuracy~\cite{zhou2016dorefa} on ImageNet dataset, a drop of 9.9\% compared to a real-valued ResNet-50. This performance gap mainly results from the quantization error in representing real-valued weights and activations with a limited number of quantized levels and the inflexibility for uniform quantizers to adapt to different distributions of input values. 

To better fit underlying distributions and mitigate quantization errors, several previous studies proposed nonuniform quantization by adjusting the quantization resolution according to the density of real-valued distribution~\cite{zhang2018lq,yamamoto2021learnable}. However, the accuracy improvement of nonuniform quantization usually comes at the expense of hardware implementation efficiency~\cite{zhou2016dorefa,gholami2021survey}. Since the output of the nonuniform quantization are floating-point weights and activations, their multiplication can no longer be directly accelerated by the bitwise operation between binaries~\cite{jeon2020biqgemm}. A common solution to address this issue has been building look-up tables (LUTs) to map floating-point values to binary digits~\cite{yamamoto2021learnable,han2015deep}, as shown in Fig.~\ref{fig:overall}(a). This post-processing incurred by nonuniform quantization costs more hardware area and consumes additional energy compared to uniform quantization~\cite{arish2015efficient,gholami2021survey}.

The goal of this study is to develop a new quantization method maintaining the hardware projection simplicity as uniform quantization and meanwhile offering the flexibility to achieve the merit on performance of nonuniform quantization.
Despite that it is desirable to enforce a quantizer's outputs (\textit{i.e.}, quantized weights/activations) to have uniform quantization levels in order not to incur additional post-processing tasks, each of the output levels does not necessarily need to represent an equal range of the real-valued input. As shown in Fig.~\ref{fig:overall} (b), we enforce the output quantization levels to be equidistant while learning the thresholds on the input values to incorporate more flexibility in fitting the underlying real-valued distributions for quantization.
We name this quantizer design as Nonuniform-to-Uniform Quantizer (N2UQ). 

However, it is challenging to optimize such a non-uniform-to-uniform quantizer, which can automatically learn to adapt the input thresholds through network training for higher precision. Because the gradient computation \textit{w.r.t.} the threshold parameters is intractable, and cannot be resolved by the existing gradient estimation method for quantization, \textit{i.e.}, the straight-through estimator (STE)~\cite{bengio2013estimating}.
STE simply estimates the incoming gradient to a threshold operation to be equal to the outgoing gradient, which by definition is unable to incorporate the threshold difference in training neither to update the threshold with gradients. 

To circumvent this challenge, we revisit the earliest derivation of STE from the stochastic binarization~\cite{bengio2013estimating} and derive a novel and more flexible backward approximation method for quantization. We name it as Generalized Straight-Through Estimator (G-STE). It degenerates to STE when all the input intervals are equal-sized, while for the scenarios that require nonuniform input thresholds, it automatically adapts the thresholds with gradient learning and provides a finer-grained approximation to the quantization function. Specifically, G-STE encodes the expectation of stochastic quantization into the backward approximation to the forward deterministic quantization functions, which naturally converts the intractable gradient computation \textit{w.r.t.} the input threshold parameters to that \textit{w.r.t.} the slopes, and encodes the influence from input threshold difference to the remaining network in the backward gradient computation.

Moreover, we propose the weight regularization that considers the overall entropy in a weight filter to further reduce the information loss arising from weight quantization.
We extensively evaluate the effectiveness of the proposed N2UQ with the collective contributions of the threshold-learning quantizer via G-STE and the weight regularization on ImageNet with different architectures and different bit-width constraints. Under all deployment scenarios, N2UQ consistently improves the accuracy by a significant margin compared to the state-of-the-art methods, including both uniform and nonuniform quantization. 

The contribution of this paper includes four aspects:

\noindent$\bullet$ We propose Nonuniform-to-Uniform Quantizer (N2UQ) for improving the quantization precision via learning input thresholds, while maintaining hardware-friendliness in implementation similar to uniform quantization.

\noindent$\bullet$ We propose Generalized Straight-Through Estimator (G-STE) to tackle intractable gradient computation \textit{w.r.t.} input threshold parameters in N2UQ. G-STE calculates the expectation of the stochastic quantization as the backward approximation to the forward deterministic quantization.

\noindent$\bullet$ Based on entropy analysis, we propose a novel weight regularization considering the overall weight distribution for further preserving information in weight quantization.

\noindent$\bullet$ We demonstrate that even under the strict constraint of fixing the quantized weights and activations to be uniform and only learning input thresholds, N2UQ exceeds state-of-the-art nonuniform quantization method with $0.5\!\!\sim\!\!1.7$\% higher accuracy on ImageNet. Specifically, the 2-bit ResNet-50 model achieves 76.4\% top-1 accuracy on ImageNet, reducing the gap to its real-valued counterpart to only 0.6\%, demonstrating the effectiveness of N2UQ design.

\section{Related Work}
Model compression is a useful technology for deploying neural network models to mobile devices with limited storage and computational power~\cite{cai2018proxylessnas,cai2019once}, and has attracted increasing attention.
Model compression methods can be categorized into several major categories, including quantization~\cite{zhuang2018towards,zhang2018lq,zhou2016dorefa,liu2018bi}, pruning~\cite{ding2019global,liu2017learning,liu2019metapruning}, knowledge distillation~\cite{hinton2015distilling,shen2019meal}, compact network design~\cite{howard2017mobilenets,sandler2018mobilenetv2,ma2018shufflenet,zhang2018shufflenet}, etc. This work is mainly focused on quantization.

Quantization can be further classified to uniform quantization~\cite{zhou2016dorefa,choi2018pact,gong2019differentiable,jacob2018quantization,hubara2017quantized,choi2019accurate} and nonuniform quantization~\cite{li2020additive,zhang2018lq,yamamoto2021learnable,miyashita2016convolutional}. Compared to uniform approaches, nonuniform quantization may achieve higher accuracy because it can better capture the underlying distributions by learning to allocate more quantization levels to important value regions~\cite{gholami2021survey}. For example, PoT~\cite{miyashita2016convolutional} uses powers-of-two levels and ApoT~\cite{li2020additive} further proposed additive powers-of-two quantization. However, it is typically difficult to deploy nonuniform quantization efficiently on the general-purpose hardware~\cite{yamamoto2021learnable}, \textit{e.g.}, GPU and CPU. Since to accelerate nonuniform quantization with bitwise operations requires additional operations or designs like look-up tables (LUTs) for mapping between floating-point quantization outputs and their binary digit representations~\cite{jeon2020biqgemm,han2015deep}, which is less efficient compared to uniform quantization.
Our proposed N2UQ is motivated to combine the merits of uniform and nonuniform quantization by outputting uniformly quantized levels for hardware-friendly implementation while allowing effective input thresholds learning to fit the underlying distributions for higher accuracy. 

Moreover, the quantization function is intrinsically a discontinuous step function and nearly always has zero gradients \textit{w.r.t.} inputs. To circumvent this problem, early work proposed straight-through estimation (STE)~\cite{bengio2013estimating} for gradient estimation, which is widely adopted in subsequent quantization methods~\cite{zhang2018lq,yamamoto2021learnable,hubara2017quantized,zhou2016dorefa,zhuang2018towards,choi2018pact,esser2020learned}. However, STE simply backpropagates through the hard threshold function as if it had been the identity function, which is unable to incorporate non-linearities in quantizer's input and output to fit various distributions. To deal with this restriction, previous works manually add non-linear functions before and/or after the STE-based quantizer. For example, LCQ~\cite{yamamoto2021learnable} adds compressing and expanding function to STE-based quantizer to achieve nonuniform quantization. QIL~\cite{jung2019learning} proposes to learn a power function before the STE-based quantizer for controlling quantization interval sizes. However, these methods relying on STE suffer from the discrepancy between quantizer's thresholds and nonlinear function's turning points~\cite{yamamoto2021learnable} or the inflexibility of using a human-defined smooth function to control all the quantization intervals with a single power hyper-parameter~\cite{jung2019learning}.
To overcome this limitation, we proposed Generalized Straight-Through Estimator (G-STE) to incorporate flexible non-linearity learning inside the quantizer, as detailed in Sec.~\ref{sec:method}.

\section{Method}
\label{sec:method}
In this section, we first briefly introduce the objective of quantization in Sec.~\ref{sec:preliminaries}. Then we present our nonuniform-to-uniform quantization (N2UQ) in Sec.~\ref{sec:N2UQ} with three components: threshold learning quantizer in the forward pass in Sec.~\ref{sec:forward}, its backward approximation with Generalized Straight-Through Estimator (G-STE) in Sec.~\ref{sec:gste} and an additional weight regularization method in Sec.~\ref{sec:weight_quan}.

\subsection{Preliminaries}
\label{sec:preliminaries}
Matrix multiplication is the most computation-costly operation in deep neural networks.
By quantizing the weights and activations in the convolutional layers and fully connected layers to fixed bits, the matrix multiplication can be accelerated with bitwise operations ~\cite{zhou2016dorefa, zhang2018lq}:
\begin{small}
\begin{equation}
\label{eq:bitwise_operation}
\begin{split}
a^q \cdot w^q = \sum_{i=0}^{M-1}\sum_{j=0}^{K-1}2^{i+j}\text{popcnt}[\text{and}(\mathbf{a}_i,\mathbf{w}_j)],&\\
\mathbf{a}_i,\!\mathbf{w}_j \!\!\in\!\! \{0, 1\} \ \forall i\!\in\!\{0, 1, ..., M\!-\!1\}, j\!\in\!\{0, 1, ..., K\!-\!1\}.&
\end{split}
\end{equation}
\end{small}

Here $\mathbf{a}$ and $\mathbf{w}$ are binary vectors of the quantized activation $a^q$ and quantized weight $w^q$. $M$ and $K$ denote the number of bits used in representing $a^q$ and $w^q$, respectively.
While Eq.~\ref{eq:bitwise_operation} in general holds true for uniform quantization, the situation becomes complicated when considering the nonuniform quantization.

\subsection{Nonuniform-to-Uniform Quantization}
\label{sec:N2UQ}

To sort out the inner mechanism of quantization, we summarize two properties of the quantizer: 

\noindent \textbf{Property 1} (Quantizer’s output). \textit{The premise of quantized network being accelerated with Eq.~\ref{eq:bitwise_operation} is that the quantizer’s output (\textit{i.e.}, quantized weights $w^q$ and activations $a^q$) can be represented in OR mapped to binary digits within the fixed bits: $ a^q = \sum_{i=0}^{M-1}\mathbf{a}_i2^i$ $, w^q = \sum_{j=0}^{K-1}\mathbf{w}_j2^j$~\cite{zhou2016dorefa}}.

This property is straightforward since Eq.~\ref{eq:bitwise_operation} takes place between binary vectors. While uniformly quantized $a^q$ and $w^q$ can be easily converted to binaries via linear mappings, it usually requires additional operations~\cite{li2020additive} or look-up tables (LUTs)~\cite{yamamoto2021learnable, han2015deep} for encoding ‘$n$-bit’ nonuniformly quantized $a^q$ and $w^q$ to $n$-bit binary digits, because outputs of $n$-bit nonuniform quantizer are actually $2^n$ in-equidistant floating-point values, as illustrated in Fig.~\ref{fig:overall}. This post-processing step incurs larger hardware area and more energy consumption as mentioned in hardware studies~\cite{arish2015efficient, jain2018compensated}.

\noindent \textbf{Property 2} (Quantizer’s input): 
\textit{The uniformness in input and output levels of the quantization function $x^q=F_Q(x^r)$ can be detached through a proper quantizer design, \textit{i.e.}, uniform quantization levels in output $x^q$ do not necessarily need to represent uniform ranges from the input $x^r$.}

This property originates from the observation that the quantizer represents real-valued variables $x^r$ with several quantized levels $x^q$, while the quantized levels are fixed, the real-valued distributions are diverse. Thus, allowing learnable thresholds in dividing $x^r$ to fixed quantized levels $x^q$ theoretically can incorporate more representational flexibility for achieving higher precision.

\subsubsection{\!\!Forward Pass: Threshold Learning Quantization}
\label{sec:forward}
Based on these two properties, we develop nonuniform-to-uniform quantizer (N2UQ) for activation quantization, with its forward pass formulated as:
\begin{align}
\label{eq:def}
x^q = \left\{ 
            \begin{array}{lr}
            0 & x^r < T_1 \\
            1 & T_1 \leqslant x^r < T_2 \\
            ...\  & ...\\
            2^n-1  & x^r \geqslant T_{2^n-1}  \\
            \end{array} 
        \right.,
\end{align}
where $n$ is the number of bits, $T$ represents learnable thresholds, and $x^r$, $x^q$ represent the input real-valued variables and the output quantized variables, respectively. The goal of N2UQ is to encode threshold learning in the quantizer to allow more flexibility, while output uniformly quantized weights and activations to accommodate fast bitwise operations without the post-processing step between quantization and matrix multiplication.

\subsubsection{Backward Pass: Generalized Straight-Through Estimator (G-STE)}
\label{sec:gste}

However, backpropagating through Eq.~\ref{eq:def} is difficult because: \textbf{(I)} The derivatives of Eq.~\ref{eq:def} \textit{w.r.t.} input $x_r$ are almost zero everywhere. \textbf{(II)} The gradient calculation \textit{w.r.t.} threshold parameters is intractable. 

For challenge \textbf{(I)}, previous quantization works adopt straight-through estimator (STE) to approximate the backward gradients in the quantization functions~\cite{zhou2016dorefa, choi2018pact, hubara2017quantized, yamamoto2021learnable, liu2018bi_journal, wu2018mixed}. STE simply assigns the incoming gradients to a threshold operation to be the outgoing gradients:
\begin{equation}
    \frac{\partial \mathcal{L}}{\partial x^q_{i,l}} = \frac{\partial \mathcal{L}}{\partial x^r_{i,l}}
\end{equation}
$x^q_{i,l}$ and $x^r_{i,l}$ denote the $i^{th}$ quantized / real-valued variables in $l^{th}$ layer, respectively.
This simple approximation function works well for uniform quantizers. However, STE implicitly enforces the equal axis aspect ratio in the input and output intervals of the quantizer because it regards the quantization function as an identity function in the backward pass. This obstacles the quantizer design from allowing learnable input thresholds while fixing the output levels. Moreover, STE cannot deal with challenge \textbf{(II)}, gradient computation \textit{w.r.t.} learnable thresholds since STE basically just bypasses the gradients as if the quantization function does not exist. Thus, to come up with a finer-grained and more flexible backward approximation that can tackle the gradient computation \textit{w.r.t.} input thresholds, we revisit STE in stochastic binarization and derive the proposed Generalized Straight-Through Estimator (G-STE).

\noindent \textbf{Lemma 1}. \textit{In binarization, straight-through estimation (STE) for gradient approximation to the forward deterministic binarization function can be derived from the expectation of the stochastic binarization function~\cite{hubara2017quantized}}.

The concept of stochastic binarization and deterministic binarization are first proposed in~\cite{bengio2013estimating, courbariaux2015binaryconnect, courbariaux2016binarized}. In stochastic binarization, real-valued variables are binarized to $-1/1$ stochastically according to their distances to $-1/1$:
\begin{equation}
\label{eq:prob_binary}
\resizebox{0.43\textwidth}{!}{$\begin{split}
\widetilde{x}^{b}_{i,l} = \left\{
             \begin{array}{lr}
             \vspace{0.5em} 
             \!-1 \!\! & \text{with probability} \ \ p = {\rm clip}( \frac{1-x^r_{i,l}}{2}, 0, 1) \\
             \!1 \!\! & \text{with probability} \ \ p = {\rm clip}( \frac{1+x^r_{i,l}}{2}, 0, 1)
             \end{array}
\right.,
\end{split}$}
\end{equation}
where $\widetilde{x}^b_{i,l}$ denotes the stochastic binary variables. To update $W_{ij,l}$ (weights in layer $l$ connecting neuron $j$ in layer $l-1$ to neuron $i$ in layer $l$), the expected gradients through the stochastic binarization function are computed: 
\begin{equation}
\mathbb{E}[\frac{\partial \mathcal{L}}{\partial W_{ij,l}}] = \mathbb{E}_{/\widetilde{x}^b_{i,l}}[\frac{\partial \mathcal{L}}{\partial \widetilde{x}^b_{i,l}}\mathbb{E}_{\widetilde{x}^b_{i,l}}[\frac{\partial \widetilde{x}^b_{i,l}}{\partial x^r_{i,l}}]\frac{\partial x^r_{i,l}}{\partial W_{ij,l}}],
\end{equation}
where $\mathbb{E}$, $\mathbb{E}_{\widetilde{x}^b_{i,l}}$ and $\mathbb{E}_{/\widetilde{x}^b_{i,l}}$ are respectively, the expectation over the whole network, only the stochastic binary variables and other parts except the stochastic binary variables. Specifically,
\begin{equation}
\begin{split}
\label{eq:back_binary}
& \mathbb{E}_{\widetilde{x}^b_{i,l}}[\frac{\partial \widetilde{x}^b_{i,l}}{\partial x^r_{i,l}}] = \frac{\partial}{\partial x^r_{i,l}} \mathbb{E}[\widetilde{x}^b_{i,l}] \\
& \ \ = \frac{\partial}{\partial x^r_{i,l}} (-1 \times p_{\{\widetilde{x}^b_{i,l} = -1\}} + 1 \times p_{\{\widetilde{x}^b_{i,l} = 1\}}) \\
& \ \ = \frac{\partial}{\partial x^r_{i,l}} ({\rm clip}(x^r_{i,l},-1,1) \\
\end{split}
\end{equation}
This arrives at the common straight-through estimator for the binarization functions~\cite{rastegari2016xnor,bethge2020meliusnet, zhu2019binary}, which transmits the gradients identically near the threshold and disregards the gradients when the real-valued inputs are too far from the threshold (\textit{i.e.}, $x^r\!\!>\!\!1$ or $x^r\!\!<\!\!-\!1$). Meanwhile, the widely-adopted deterministic binarization function in forward pass~\cite{liu2018bi,liu2018bi_journal,liu2020reactnet,bulat2019xnor,courbariaux2015binaryconnect} can be attained via setting a hard threshold on the probability (\textit{i.e.}, $p=0.5$) in Eq.~\ref{eq:prob_binary}:
\begin{equation}
\begin{split}
x^b_{i,l} = \left\{  
             \begin{array}{lr} 
             \!\!-1 &   p_{\{\widetilde{x}^b_{i,l} = -1\}} > 0.5 \\
             \!\!1 &  p_{\{\widetilde{x}^b_{i,l} = -1\}} \leqslant 0.5 
             \end{array}  
\right. \!\! = \left\{  
             \begin{array}{lr}  
             \!\!-1 & \!x^r_{i,l} < 0 \\ 
             \!\!1 & \! x^r_{i,l} \geqslant 0  
             \end{array}  
\right.
\end{split}
\end{equation}
To this end, we show that STE encodes the expectation of stochastic binarization in the backward approximation to the forward deterministic binarization functions. 

\begin{figure}[t]
\centering
\includegraphics[width=\linewidth]{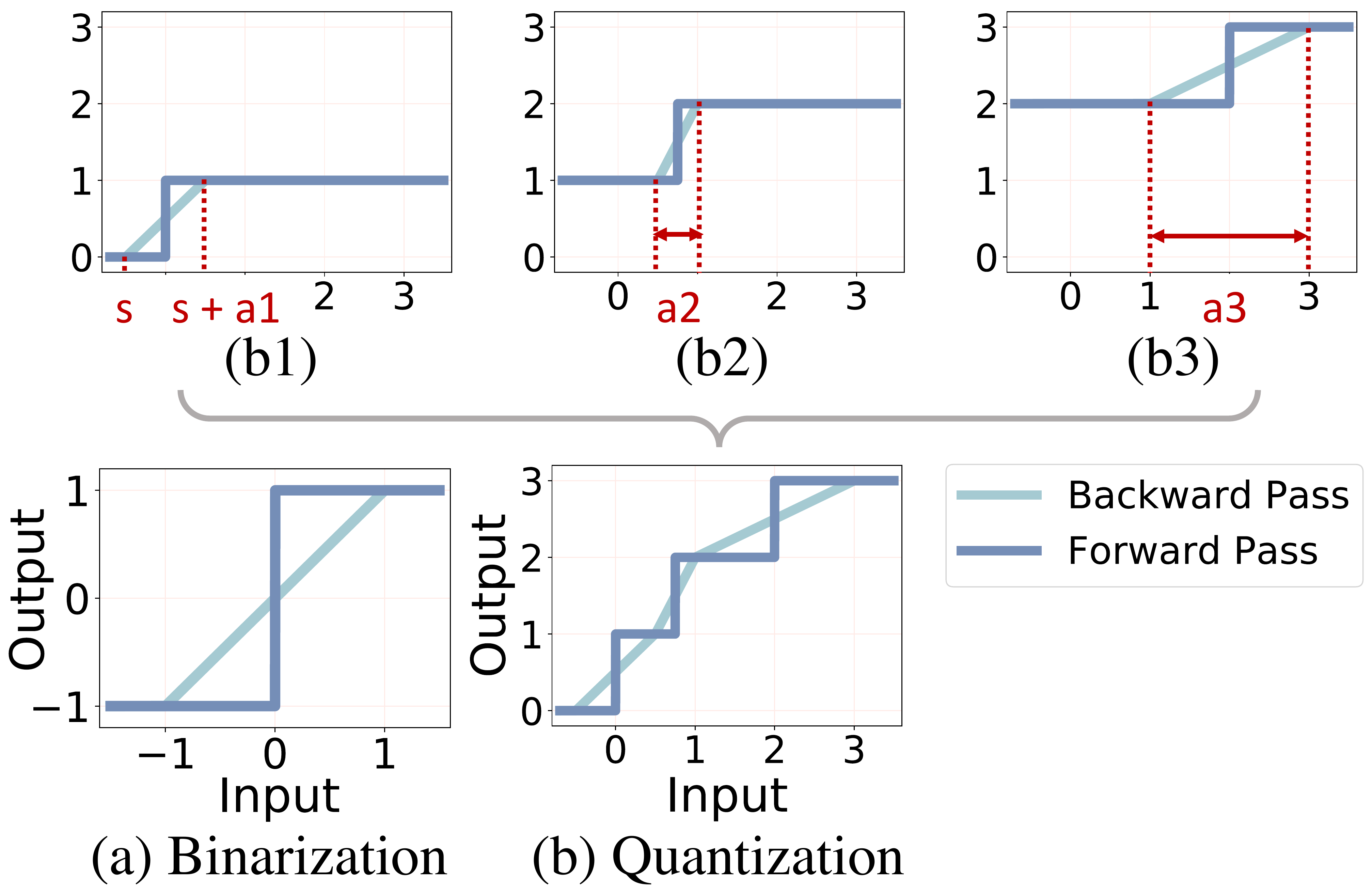}
\vspace{-1em}
\caption{(a) The deterministic binarization function and straight-through estimation (STE) as the backward approximation. (b) Quantization can be viewed as a summation of segments of binarization functions, which derives the proposed generalized straight-through estimator (G-STE).}
\label{fig:quan_derive}
\end{figure}

\noindent \textbf{Lemma 2}. \textit{A quantization function $x^q$ can be regarded as a summation of binarization functions $x^b$ with different thresholds: $x^q = x^b_{thre_1} + x^b_{thre_2} + ... + x^b_{thre_n}$.}

This lemma is self-evident, as illustrated in Fig.~\ref{fig:quan_derive} (b). From here, we extend the concept of stochastic binarization to quantization for deriving G-STE to solve the threshold learning problem in nonuniform-to-uniform quantizer.

We start with the first quantization segment. We denote the initial point as $s$ and its length as $a_1$. For simplification, we omit the subscripts here. Following the concept of stochastic binarization, in the range of $[s, s+a_1]$, real-valued variables can be quantized to $0/1$ stochastically with the probability proportional to their distance to $s/s+a_1$:
\begin{equation}
\label{eq:forward_quan_seg1}
\resizebox{0.43\textwidth}{!}{$\begin{split}
\widetilde{x}^{q_{\{0,1\}}} = \left\{  
             \begin{array}{lr}
             \vspace{0.5em} 
             \!0 \ \ \ \text{with probability} \ p\!=\!{\rm clip}(\frac{s + a_1 - x^r}{a_1}, 0, 1) \\
             \!1 \ \ \ \text{with probability} \ p\!=\!{\rm clip}( \frac{x^r-s}{a_1}, 0, 1)
             \end{array}  
\right. ,
\end{split}$}
\end{equation}
$\widetilde{x}^{q_{\{0,1\}}}$ is the stochastic quantization output within this segment. 
Similar to Eq.~\ref{eq:back_binary}, the derivation of this quantization segment can be computed from the expectation of Eq.~\ref{eq:forward_quan_seg1}:
\begin{equation}
\begin{split}
\label{eq:back_a1}
&\mathbb{E}[\frac{\partial \widetilde{x}^{q_{\{0,1\}}}}{\partial x^r}] = 
\frac{\partial}{\partial x^r} \mathbb{E}[\widetilde{x}^{q_{\{0,1\}}}] \\
& \ \ \ = \frac{\partial}{\partial x^r} (0 \times p_{\{\widetilde{x}^q = 0\}} + 1 \times p_{\{\widetilde{x}^q = 1\}}) \\
& \ \ \ = \frac{\partial {\rm clip}(\frac{x^r-s}{a_1}, 0, 1)}{\partial x^r}.
\end{split}
\end{equation}
In this way, the influence of the threshold parameter $a_1$ to the network is decently encoded in the backward approximation function Eq.~\ref{eq:back_a1} by altering the slopes. Further in the forward pass, instead of using stochastic quantization which requires random seeds generating process~\cite{torii2016asic}, deterministic quantization are adopted and derived by setting the probability threshold to be $p=0.5$, 
\begin{equation}
\resizebox{0.423\textwidth}{!}{$\begin{split}
x^{q_{\{0,1\}}} = \left\{  
             \begin{array}{lr} 
             \vspace{0.2em} 
             \!\!0 &   p_{\{\widetilde{x}^q = 0\}} > 0.5 \\
             \!\!1 &  p_{\{\widetilde{x}^q = 0\}} \leqslant 0.5 
             \end{array}  
\right. \!= \left\{  
             \begin{array}{lr}
             \vspace{0.2em} 
             \!\!0 &  x^r \! < \! s + \frac{a_1}{2}\\
             \!\!1 &  x^r \! \geqslant \! s + \frac{a_1}{2}
             \end{array}  
\right. \!\!.
\end{split}$}
\end{equation}
Likewise, the forward deterministic quantization functions and the corresponding backward approximation functions for the remaining segments can be obtained from this probabilistic sense. Then, the nonuniform-to-uniform quantizer with its backward approximation for $n$-bit quantization can be naturally derived by summing up $2^n\!-\!1$ segments, as illustrated in Fig.~\ref{fig:quan_derive}. We name this proposed backward gradient approximation method as Generalized Straight-Through Estimator (G-STE) and obtain the following theorem.

\noindent \textbf{Theorem 1} \textit{Generalized straight-through estimator:}
\begin{equation}
\begin{split}
\label{eq:back_all}
&\frac{\partial x^q}{\partial x^r} = \mathbb{E}[\frac{\partial \widetilde{x}^q}{\partial x^r}] = 
\frac{\partial}{\partial x^r} \mathbb{E}[\widetilde{x}^q] \\
\vspace{0.5em} 
& \ \ \ = \left\{  
        \begin{array}{lr}
        \vspace{-0.4em} 
        \!\! \frac{\partial}{\partial x^r}(\frac{x^r-d_{i-1}}{a_i} + i-1) &  d_{i-1} \leqslant x^r < d_i \\
        \vspace{0.5em} 
        &  _{i \in \{1,..., 2^n -1\}}\\
        \!\! 0 &  otherwise
        \end{array}  
\right. ,
\end{split}
\end{equation}
\textit{as the backward gradient approximation function to an $n$-bit nonuniform-to-uniform quantizer:}
\begin{equation}
\label{eq:forward_all}
\begin{split}
x^q = \left\{  
             \begin{array}{lr}
             \vspace{0.5em} 
             \!\!0 &  x^r \! < \! d_0 + \frac{a_1}{2}\\
             \vspace{-0.2em} 
             \!\!i &  d_{i-1} + \frac{a_{i}}{2} \leqslant x^r \! < \! d_i + \frac{a_{i+1}}{2}\\
             \vspace{0.5em} 
             &  _{i \in \{1,..., 2^n -2\}}\\
             \!\!2^n-1 &  x^r \geqslant d_{2^n-2} + \frac{a_{2^n-1}}{2}
             \end{array}  
\right. .
\end{split}
\end{equation}
\textit{tackles the threshold learning problem by using the expectation of the stochastic version of the nonuniform-to-uniform quantizer as its backward approximation. Here $d_0 = s, \ \ d_i = s + \sum_{j=1}^{i} a_j, \ \ i\in\{1,...,2^n -1\}$.} 

The essence of the nonuniform-to-uniform quantizer with G-STE backward approximation is to encode the mean of stochastic quantization in the backward approximation function while having deterministic quantization function with hard thresholds in forward pass. This helps assimilate the influence of threshold parameters of forcing outputs into particular values in the forward pass into approximating the squeezed or expanded probability of quantization in the backward pass, in which the influence of the threshold parameters $a_i$ to the remaining neural networks is reflected in the slopes of the backward function. In turn the gradients \textit{w.r.t.} threshold parameters can be easily computed as,
\begin{equation}
\begin{split}
\hspace{-0.6em}\frac{\partial x^q}{\partial a_i} = \frac{\partial\mathbb{E}[x^q]}{\partial a_i} = \left\{  
        \begin{array}{lr}
        \vspace{0.5em} 
        \!\! - \frac{x^r-d_{i-1}}{a_i^2} &  d_{i-1} \leqslant x^r < d_i \\
        \vspace{-0.4em} 
        \!\! - \frac{1}{a_j}  & d_{j-1} \leqslant x^r < d_j \\
        \vspace{0.5em} 
        &_{j \in \{i+1,..., 2^n -1\}}\\
        \!\! 0 & otherwise \\
        \end{array}  
\right. .
\end{split}
\end{equation}

We can see that, when all the intervals are equal-sized, G-STE degenerates to STE, while during training it learns to automatically adjust the input thresholds with the gradients calculated from the network.
Specifically, the output levels are scaled to $\{0,2/(2^n\!-\!1), …,2\}$. We additionally apply two learnable scaling parameters $\beta_1, \beta_2$ for scaling the input $x^r$ before the quantization: $x^r := x^r \times \beta_1$ and the output $x^q$ after the quantization: $x^q := x^q \times \beta_2$. $\beta_1, \beta_2$ are initialized to 1, $a_i$ are initialized to $2/(2^n\!-\!1)$ and enforced to be positive. The proposed nonuniform-to-uniform quantizer introduces only $2^n + 2$ extra parameters per layer compared to a classical uniform quantizer, which is negligible considering the large number of network weights. 

\subsubsection{Entropy Preserving Weight Regularization}
\label{sec:weight_quan}
Further, we propose weight regularization to encourage more information-carrying capacity in quantized weights. An important observation is that weights in real-valued networks are usually small in magnitude, \textit{e.g,} $\sim\!\!10^{-2}$ for weights in a pre-trained ResNet-18 network, but the quantized weights usually expand in the range of $ [-1, 1]$. This mismatch in magnitude will cause quantized weights to collapse to a few quantization levels close to zero. The classic solution is using a $tanh$ function and dividing the maximum absolute weight value to rescale weights to $[-1, 1]$~\cite{zhou2016dorefa}. However, these rescaled weights are likely to be dominant by a few extrema and still not fully occupy the range of $[-1, 1]$, as will further be illustrated in Fig.~\ref{fig:hist_visualization}, which causes a huge information loss.

From the perspective of information theory, more information are preserved when quantized weights contain higher entropy. Thus, we are motivated to regularize the real-valued weights before the quantizer to obtain maximal entropy in quantized weights:
$\max \ \mathcal{H} = -p_i \log(p_i), s.t. \sum_{i=1}^{N} p_i = 1 $. Here $p_i$ is the proportion of real-valued weights being quantized to $i^{th}$ quantization level and $N$ is the number of quantization levels in total.
Based on the Lagrange multiplier, it is easy to obtain the optimal $p_i^* = \frac{1}{N}, \ \ i \in \{1,2,…, N\}$, meaning that, when the proportions of real-valued weights being quantized to multiple quantization levels are equal, the information entropy in the quantized weights reaches its maximum value.

Given the quantization function $F_{Q}\!=\!round(({\rm clip}($ $w^{r'},\!-\!1, 1)\!+\!1)\!\times\!\frac{2^n\!-\!\!1}{2})\!\times\!\frac{2}{2^n\!-\!\!1}\!-\!1$, we empirically solve that when real-valued weights are normalized to $W^{r'} = \frac{2^{(n-1)}}{2^n-1} \frac{|W^r|}{||W^r||_{l1}}W^r$,	
the corresponding quantized weights $w^q\!=\! F_{Q}(w^{r’},n)$ are approximately uniformly distributed in all quantization levels. Here, $W^r$ is the real-valued weight filter, $|W^r|$ denotes the number of entries in $W^r$. Different from activations that are generated from images and vary from batch to batch, weights are static parameters that have high flexibility to adjust each of their values. Therefore, instead of learning the threshold parameters to determine all the quantized weights values, we adopt equidistant thresholds for weight quantization, letting each weight individually learn its value and scale the weights based on the overall statistics. More details of the derivation can be found in Appendix.

\section{Experiments}
To verify the effectiveness of the proposed Nonuniform-to-Uniform Quantization (N2UQ), we conduct experiments on the ImageNet dataset. We first introduce the dataset and training strategy in Sec.~\ref{sec:exp_setting}, followed by the comparison to state-of-the-art quantization methods in Sec.~\ref{sec:sota}. We then analyze the effect of each proposed component of N2UQ in Sec.~\ref{sec:ablation}. Visualization results on how N2UQ captures fine-grained underlying distributions are presented in Sec.~\ref{sec:visualization}.

\subsection{Experimental Settings}
\label{sec:exp_setting}

\noindent\textbf{Dataset}
The experiments are carried out on the ImageNet-2012 classification dataset~\cite{imagenet}, with $1.2$ million training
images and $50,000$ validation images in $1000$ classes.

\noindent\textbf{Training Strategy}
We follow the training scheme in~\cite{zhou2016dorefa, yamamoto2021learnable, jung2019learning, esser2020learned} to use real-valued PyTorch pre-trained models as initialization for corresponding quantized networks. 
We use Adam optimizer~\cite{kingma2014adam} with a linear learning rate decay scheduler. The initial learning rate is set to $2.5e\!-\!3$ for weight parameters and batch size is set to $512$. We set weight decay to be 0 following~\cite{liu2018bi,liu2020reactnet}. The models are trained for $128$ epochs with the same knowledge distillation scheme as LSQ~\cite{esser2020learned}. We adopt the basic data augmentation as ResNet~\cite{he2016deep}. Training images are randomly resized and cropped to $224\times224$ pixels and randomly flipped horizontally. Test images are center-cropped to $224\times224$ pixels. For fair comparison, the floating-point (FP) model results reported in Table~\ref{table:SOTA_resnet} are initialized and fine-tuned with the same settings as quantized models.

\noindent\textbf{Learnable Parameters}
For the learnable parameters, $s$ are initialized to $0$, $a_i$ are initialized to $2/(2^n\!-\!1)$ and enforced to be greater than $1e\!-\!3$. The learnable scaling parameters $\beta_1, \beta_2$ are initialized to $1$. All these learnable parameters use $1/10$ of the learning rate as that for weight parameters. 

\noindent\textbf{Network Structure} We adopt Pre-Activation structures (\textit{i.e.}, NonLinear-Conv-BN structure) and use RPReLU~\cite{liu2020reactnet} as non-linear function. All convolution and fully-connected layers are quantized with N2UQ except for the first and the
last one.

\setlength{\tabcolsep}{2pt}
\begin{table*}[t]
\begin{center}
\caption{Accuracy comparison to the state-of-the-art quantization methods with ResNet structure on ImageNet dataset. Note that W/A denote the bit-width of weights and activations, respectively. FP denotes the top-1 accuracy of the full-precision models.}
\vspace{-0.5em}
\label{table:SOTA_resnet}
\resizebox{0.72\textwidth}{!}{
\begin{tabular}{cccccccccccc}
\hline
\hline
\noalign{\smallskip}
\multirow{2}{*}{Network} & \multirow{2}{*}{Method} & Bit-width & \multicolumn{2}{c}{Accuracy(\%)} & Bit-width & \multicolumn{2}{c}{Accuracy(\%)} & Bit-width & \multicolumn{2}{c}{Accuracy(\%)} \\
& & (W/A) & Top-1 & Top-5 & (W/A)& Top-1 & Top-5 & (W/A) & Top-1 & Top-5 \\
\noalign{\smallskip}
\hline
\multirowcell{12}{ResNet-18 \\ (FP: 71.8)}
& PACT~\cite{choi2018pact} & 2/2 & 64.4 & 85.6 & 3/3 & 68.1 & 88.2 & 4/4 & 69.2 & 89.0\\ 
& DoReFa-Net~\cite{zhou2016dorefa} & 2/2 & 64.7 & 84.4 & 3/3 & 67.5 & 87.6 & 4/4 & 68.1 & 88.1\\
& LQ-Nets~\cite{zhang2018lq} & 2/2 & 64.9 & 85.9 & 3/3 & 68.2 & 87.9 & 4/4 & 69.3 & 88.8\\
& DSQ~\cite{gong2019differentiable} & 2/2 & 65.2 & -- & 3/3 & 68.7 & -- & 4/4 & 69.6 & 88.9\\
& FAQ~\cite{mckinstry2019discovering} & -- & --& -- & -- & -- & -- & 4/4 & 69.8 & 89.1\\
& QIL~\cite{jung2019learning} & 2/2 & 65.7 & -- & 3/3 & 69.2 & -- & 4/4 & 70.1 & --\\
& DAQ~\cite{lee2021distance} & 2/2 & 66.9 & -- & 3/3 & 69.6 & -- & 4/4 & 70.5 & --\\
& DNAS~\cite{wu2018mixed} & -- & --& -- & -- & -- & --& $\sim$4/$\sim$4 & 70.6 & -- \\
& APoT~\cite{li2020additive} & 2/2 & 67.3 & 87.5 & 3/3 & 69.9 & 89.2 & 4/4 & 70.7 & 89.6 \\
& LSQ~\cite{esser2020learned} & 2/2 & 67.6 & 87.6 & 3/3 & 70.2 & 89.4 & 4/4 & 71.1 & 90.0 \\
& LCQ~\cite{yamamoto2021learnable} & 2/2 & 68.9 & -- & 3/3 & 70.6 & -- & 4/4 & 71.5 & -- \\
\cline{2-11}
& \textbf{N2UQ (Ours)} & 2/2 & \textbf{69.4} & \textbf{88.4} & 3/3 & \textbf{71.9} & \textbf{90.5} & 4/4 & \textbf{72.9} & \textbf{90.9}\\
\hline
\multirowcell{9}{ResNet-34 \\ (FP: 74.9)}
& LQ-Nets~\cite{zhang2018lq} & 2/2 & 69.8 & 89.1 & 3/3 & 71.9 & 90.2 & -- & -- & -- \\
& DSQ~\cite{gong2019differentiable} & 2/2 & 70.0 & -- & 3/3 & 72.5 & -- & 4/4 & 72.8 & --\\
& FAQ~\cite{mckinstry2019discovering} & -- & --& -- & -- & -- & -- & 4/4& 73.3 & 91.3\\
& QIL~\cite{jung2019learning} & 2/2 & 70.6 & -- & 3/3 & 73.1 & -- & 4/4 & 73.7 & --\\
& APoT~\cite{li2020additive} & 2/2 & 70.9 & 89.7 & 3/3 & 73.4 & 91.1 & 4/4 & 73.8 & 91.6\\
& DAQ~\cite{lee2021distance} & 2/2 & 71.0 & -- & 3/3 & 73.1 & -- & 4/4 & 73.7 & --\\
& DNAS~\cite{wu2018mixed} & -- & --& -- & -- & -- & --& $\sim$4/$\sim$4 & 74.0 & --\\
& LSQ~\cite{esser2020learned} & 2/2 & 71.6 & 90.3 & 3/3 & 73.4 & 91.4 & 4/4 & 74.1 & 91.7\\
& LCQ~\cite{yamamoto2021learnable} & 2/2 & 72.7 & -- & 3/3 & 74.0 & -- & 4/4 & 74.3 & --\\
\cline{2-11}
& \textbf{N2UQ (Ours)} & 2/2 & \textbf{73.3} & \textbf{91.2} & 3/3 & \textbf{75.2} & \textbf{92.3} & 4/4 & \textbf{76.0} & \textbf{92.8}\\
\hline
\multirowcell{9}{ResNet-50 \\ (FP: 77.0)}
& DoReFa-Net~\cite{zhou2016dorefa} & 2/2 & 67.1 & 87.3 & 3/3 & 69.9 & 89.2 & 4/4 & 71.4 & 89.8\\
& LQ-Nets~\cite{zhang2018lq} & 2/2 & 71.5 & 90.3 & 3/3 & 74.2 & 91.6 & 4/4 & 75.1 & 92.4\\
& FAQ~\cite{mckinstry2019discovering} & -- & -- & -- & -- & -- & -- & 4/4& 76.3 & 93.0\\
& PACT~\cite{choi2018pact} & 2/2 & 72.2 & 90.5 & 3/3 & 75.3 & 92.6 & 4/4 & 76.5 & 93.2\\
& APoT~\cite{li2020additive} & 2/2 & 73.4 & 91.4 & 3/3 & 75.8 & 92.7 & 4/4 & 76.6 & 93.1\\
& LSQ~\cite{esser2020learned} & 2/2 & 73.7 & 91.5 & 3/3 & 75.8 & 92.7 & 4/4 & 76.7 & 93.2\\
& Auxi~\cite{zhuang2020training} & 2/2 & 73.8 & 91.4 & 3/3 & 75.4 & 92.4 & -- & -- & --\\
& LCQ~\cite{yamamoto2021learnable} & 2/2 & 75.1 & -- & 3/3 & 76.3 & -- & 4/4 & 76.6 & --\\
\cline{2-11}
& \textbf{N2UQ (Ours)} & 2/2 & \textbf{75.8} & \textbf{92.3} & 3/3 & \textbf{77.5} & \textbf{93.6} & 4/4 & \textbf{78.0} & \textbf{93.9}\\
\hline
\hline
\end{tabular}
}
\vspace{-0.5em}
\end{center}
\end{table*}

\setlength{\tabcolsep}{2pt}
\begin{table}[t]
\begin{center}
\caption{Accuracy comparison with MobileNet structure.}
\vspace{-0.5em}
\label{table:SOTA_mbv2}
\resizebox{0.45\textwidth}{!}{
\begin{tabular}{cccccccccccc}
\hline
\hline
\noalign{\smallskip}
Network & Method & Top1 Acc (\%) & Top5 Acc (\%) \\
\noalign{\smallskip}
\hline
\multirowcell{5}{MobileNet-V2 \\ (FP: 72.0)} & DSQ~\cite{gong2019differentiable} & 64.8 & -- \\
& LLSQ~\cite{zhao2020linear} & 67.4 & 88.0 \\ 
& LCQ~\cite{yamamoto2021learnable} & 70.8 & 89.7 \\
& PROFIT~\cite{park2020profit} & 71.6 & 90.4 \\
\cline{2-4}
& \textbf{N2UQ(Ours)} & \textbf{72.1} & \textbf{90.6}\\
\hline
\hline
\end{tabular}
}
\vspace{-1em}
\end{center}
\end{table}

\subsection{Comparison with State-of-the-Art Methods}
\label{sec:sota}

Table~\ref{table:SOTA_resnet} summarizes the accuracy of the proposed N2UQ on ResNet. Compared to uniform quantization, N2UQ allows more flexibility in learning input thresholds, and the results demonstrate that N2UQ surpasses uniform methods~\cite{esser2020learned, gong2019differentiable, zhou2016dorefa, choi2018pact} by a large margin. Moreover, the accuracy improvements become more significant with larger bit-width, as more flexibility can be incorporated with threshold learning. For nonuniform quantization, the state-of-the-art method LCQ~\cite{yamamoto2021learnable} intrinsically uses the same quantizer as the classic uniform quantizer~\cite{zhou2016dorefa, choi2018pact} and learns two additional non-linear functions to reshape the real-valued input and quantized output. Instead, the proposed N2UQ completely integrates the non-linearity learning inside the quantizer, which avoids the discrepancy between the non-linear functions' turning points and the quantizers' thresholds. Thus, the proposed N2UQ achieves up to 1.7\% higher accuracy than LCQ, showing that N2UQ is well designed to incorporate threshold learning into the quantizers for higher flexibility and better performance. Moreover, compared to nonuniform quantization, N2UQ not only saves the parameters by fixing the output levels, but also results in uniformly quantized weights and activations for computing the quantized matrix multiplication, which prevents the post-processing step of building look-up tables (LUTs) and mapping the floating-point levels to binaries, thus is much more hardware-friendly ~\cite{gholami2021survey,jacob2018quantization,jeon2020biqgemm}. For the MobileNetV2 structure, N2UQ also achieves the state-of-the-art result as shown in Table~\ref{table:SOTA_mbv2}. It is remarkable that N2UQ can compress the real-valued network to 4-bit without compromising or even improving the accuracy, indicating that deep neural network structures have redundancy and N2UQ is well-designed to fit underlying distributions and act as a good regularization to prevent overfitting as well~\cite{han2015deep}.

\subsection{Ablation Study}
\label{sec:ablation}

To investigate the effect of each component on the final performance, we conduct thorough ablation studies on the 2-bit quantized ResNet-18 network on ImageNet.
As shown in Table~\ref{table:ablation_a_and_w}, the proposed weight regularization and threshold learning quantizer with G-STE improve the accuracy by 1.9\% and 3.0\%, respectively over the baseline~\cite{zhou2016dorefa}. By using both techniques, our proposed N2UQ achieves 69.7\% accuracy, narrowing the accuracy gap to the corresponding real-valued ResNet-18 to only 2.1\%.

\begin{figure*}[t]
\centering
\includegraphics[width=0.9\linewidth]{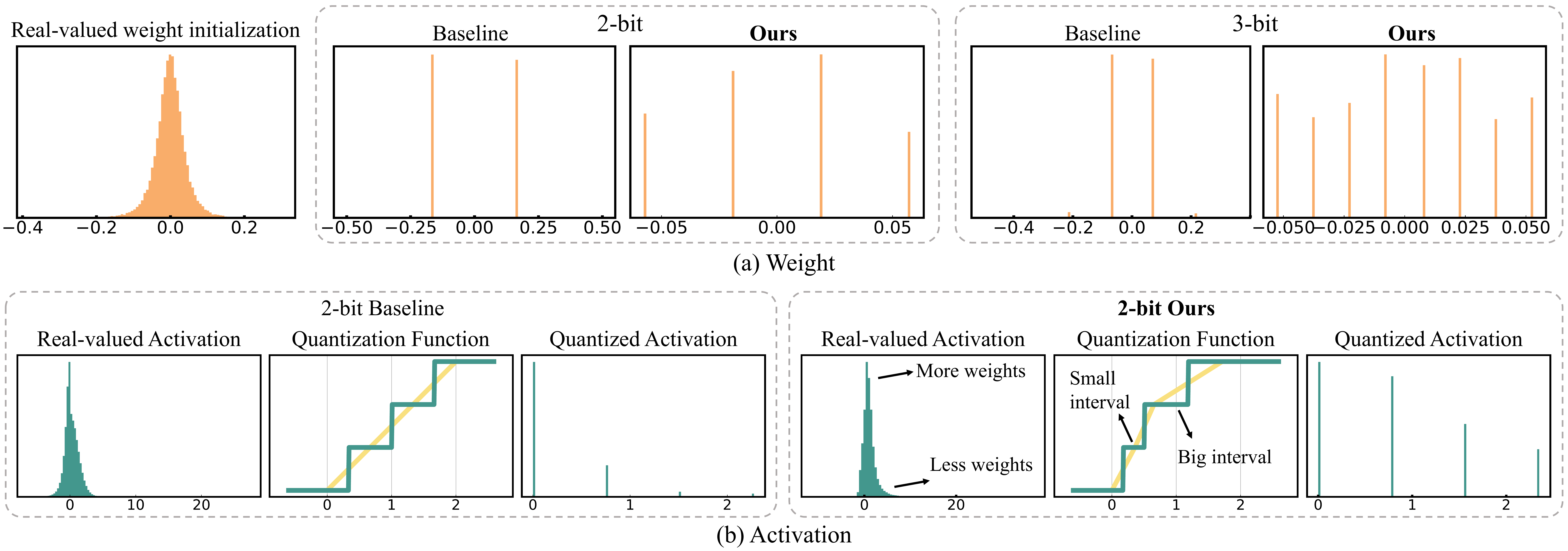}
\vspace{-0.5em}
\caption{Histogram of the weight and activation distribution.} 
\vspace{-0.5em}
\label{fig:hist_visualization}
\end{figure*}

In Table~\ref{table:ablation_w}, based on the proposed threshold learning quantizer with G-STE, we make further ablation study on weight regularization. Compared to the proposed method of calculating the regularization factor from the overall weight statistics, learning regularization factors with gradients introduces unstableness since gradients are noisy, which affects the training negatively, as shown in $4^{th}$ row of Table~\ref{table:ablation_w}. Moreover, the previous weight norm~\cite{salimans2016weight} designed for training real-valued network is not aware of weight quantization, but only scales the weights participating in matrix multiplication, which turns out unsuitable for quantized network, as shown in $3^{rd}$ row of Table~\ref{table:ablation_w}. In comparison, the entropy-based regularization considers the knowledge of weight distribution, such that it can better reduce information loss in weight quantization for higher accuracy.

\setlength{\tabcolsep}{2pt}
\begin{table}[t]
\begin{center}
\caption{Effects of different components of N2UQ on the final performance of a 2-bit quantized ResNet-18 network.}
\vspace{-1em}
\label{table:ablation_a_and_w}
\resizebox{0.48\textwidth}{!}{
\begin{tabular}{cccccc}
\noalign{\smallskip}
\hline\noalign{\smallskip}
\hline
 Method & \hspace{-1em} Top-1 Acc \\
\hline
Baseline~\cite{zhou2016dorefa} (our implementation) & 65.9 \\
+ Threshold Learning Activation Quantizer with G-STE & 68.9 \\
+ Entropy Preserving Weight Regularization & 67.8\\
+ Both (N2UQ) & \textbf{69.7}\\ 
\hline
Corresponding real-valued network & 71.8\\ 
\hline
\hline
\end{tabular}} 
\vspace{-0.5em}
\end{center}
\end{table}

\subsection{Visualization}
\label{sec:visualization} To further understand the proposed N2UQ, we illustrate the initial weight distributions and the learned parameters in quantized networks. In Fig.~\ref{fig:hist_visualization} (a), given the real-valued pre-trained weights as initialization, the baseline method~\cite{zhou2016dorefa} of using $tanh$ function and dividing the maximum weight value to normalize the weights is likely to be biased by the extrema. Specifically, for the 2-bit case, only $6$ entries in the weight filter are quantized to $0.5$ and $20$ entries are quantized to $-0.5$ in the baseline method. In contrast, the proposed entropy-based regularization method produces more evenly distributed quantized weights, which evidently enhances the amount of information that quantized weights carry.
In Fig.~\ref{fig:hist_visualization} (b), the classic uniform quantization has fixed input thresholds, making it hard to adjust itself to fit different underlying distributions. In comparison, N2UQ learns the input thresholds with neural network training. As shown in the right dashed box of Fig.~\ref{fig:hist_visualization} (b), when the real-valued distribution is dense around zero and sparse in the tail, the quantization function learns smaller intervals for the dense area and larger intervals for the sparse part, which better represents the distribution and mitigates the quantization error. 

\setlength{\tabcolsep}{2pt}
\begin{table}[t]
\begin{center}
\caption{Comparison among different weight regularization schemes for a 2-bit quantized ResNet-18 on ImageNet, based on the proposed threshold learning quantizer with G-STE.}
\vspace{-0.5em}
\label{table:ablation_w}
\resizebox{0.4\textwidth}{!}{
\begin{tabular}{cccccc}
\hline\noalign{\smallskip}
\hline
Method & Top-1 Acc \\
\hline
No Regularization & 68.9 \\ 
Weight Norm~\cite{salimans2016weight} & 67.4\\ 
Learnable Scaling Factor & 68.6 \\
Entropy Preserving Weight Regularization & \textbf{69.7}\\
\hline
\hline
\end{tabular}}
\vspace{-1em}
\end{center}
\end{table}

\section{Conclusions}
\label{sec:conclusion}
We have introduced Nonuniform-to-Uniform Quantization, abbreviated as N2UQ, for learning input thresholds in the uniform quantizer. Compared to nonuniform quantization, N2UQ enjoys the advantage of hardware-friendliness since the uniform outputs do not require additional post-processing before the bitwise operations. Meanwhile, N2UQ contains stronger representational capability than the classical uniform quantization by making the input thresholds learnable. To tackle the intractable gradient computation \textit{w.r.t.} input thresholds, we derive the generalized straight-through estimator (G-STE) from the essence of stochastic quantization. Further, we propose entropy preserving weight regularization to further reduce quantization error. With these contributions, N2UQ achieves $0.8\!\!\!\sim\!\!\!1.7$\% higher accuracy than previous state-of-the-art nonuniform methods under the harsh condition of having uniform output quantized levels. Particularly, the 2-bit N2UQ ResNet-50 is just 0.6\% shy of its real-valued counterpart.

\section*{Acknowledgments}
This research was partially supported by ACCESS - AI Chip Center for Emerging Smart Systems, sponsored by InnoHK funding, Hong Kong SAR. We would like to thank Turing AI Computing Cloud~\cite{tacc} and HKUST iSING Lab for providing us computation resources on their platform.

{\small
\bibliographystyle{ieee_fullname}
\bibliography{egbib}
}

\clearpage
\appendix

\section*{\Large{Appendix}}

\noindent In this appendix, we provide details omitted in main text:

\noindent$\bullet$ Section~\ref{sec:weight_details}: Illustration and more details about the proposed Entropy Preserving Weight Regularization.

\noindent$\bullet$ Section~\ref{sec:weight_visualization}: Visualization of learned parameters.


\noindent$\bullet$ Section~\ref{sec:resnet_more}: The results of keeping down-sampling layers to be real-valued in ResNet structures.

\section{Entropy Preserving Weight Regularization}
\label{sec:weight_details}

\begin{figure}[h]
\centering
\includegraphics[width=\linewidth]{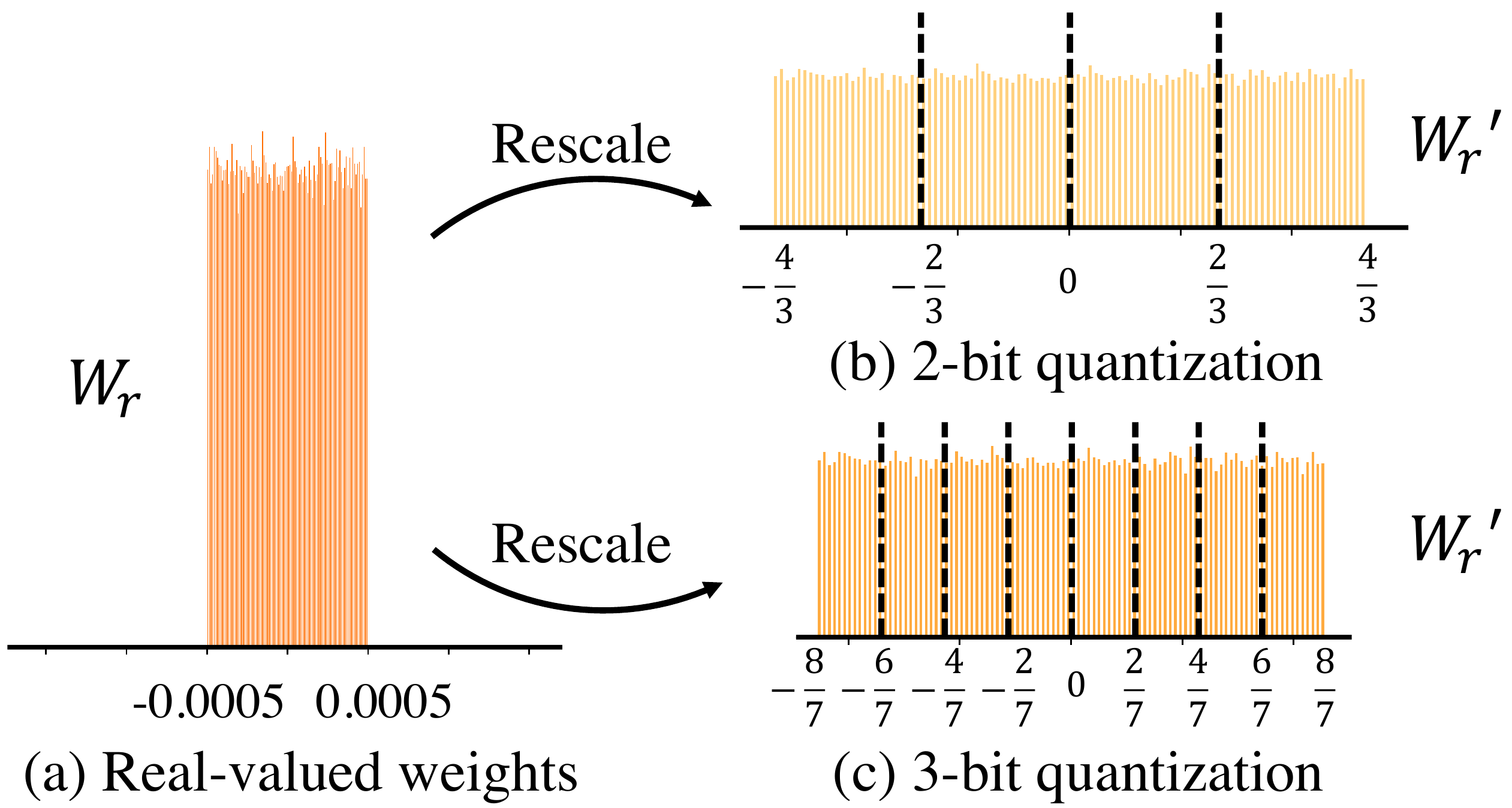}
\vspace{-1.5em}
\caption{Illustration of the proposed entropy preserving weight regularization.} 
\label{fig:weight_quan}
\end{figure}

The weight regularization function proposed in the main paper:
$W^{r'} = \frac{2^{(n-1)}}{2^n-1} \frac{|W^r|}{||W^r||_{l1}}W^r$
aims to rescale the real-valued weights for preserving information entropy in the corresponding quantized weights.
Specifically, $\frac{|W^r|}{||W^r||_{l1}}$ scales the $W^r$ to have the absolute mean value equal to 1. When the real-valued weights are initialized as uniformly and symmetrically distributed~\cite{he2015delving,glorot2010understanding}, $\frac{|W^r|}{||W^r||_{l1}}W^r$ will be evenly distributed in $[-2, 2]$.
The factor $\frac{2^{(n-1)}}{2^n-1}$ further spread the real-valued weight distribution to $[-\frac{2^n}{2^n-1}$, $\frac{2^n}{2^n-1}]$, for which, the corresponding quantized weights after the quantization function $F_{Q}\!=\!round((Clip(\!-\!1,$ $W^{r'}, 1)\!+\!1)\!\times\!\frac{2^n\!-\!\!1}{2})\!\times\!\frac{2}{2^n\!-\!\!1}\!-\!1$ will be approximately uniformly quantized to $2^n$ levels as shown in Fig.~\ref{fig:weight_quan}. During training, the real-valued weight distributions are not always uniform, in which case, regularization helps to better distribute the weights.
After training, this regularization factor can be calculated offline from the optimized weights and be absorbed by the BatchNorm layers (if used) after the quantized convolutional layers as mentioned in~\cite{liu2018bi_journal}.

\section{Learned Parameters Visualization}
\label{sec:weight_visualization}

\begin{figure}[h]
\centering
\includegraphics[width=0.9\linewidth]{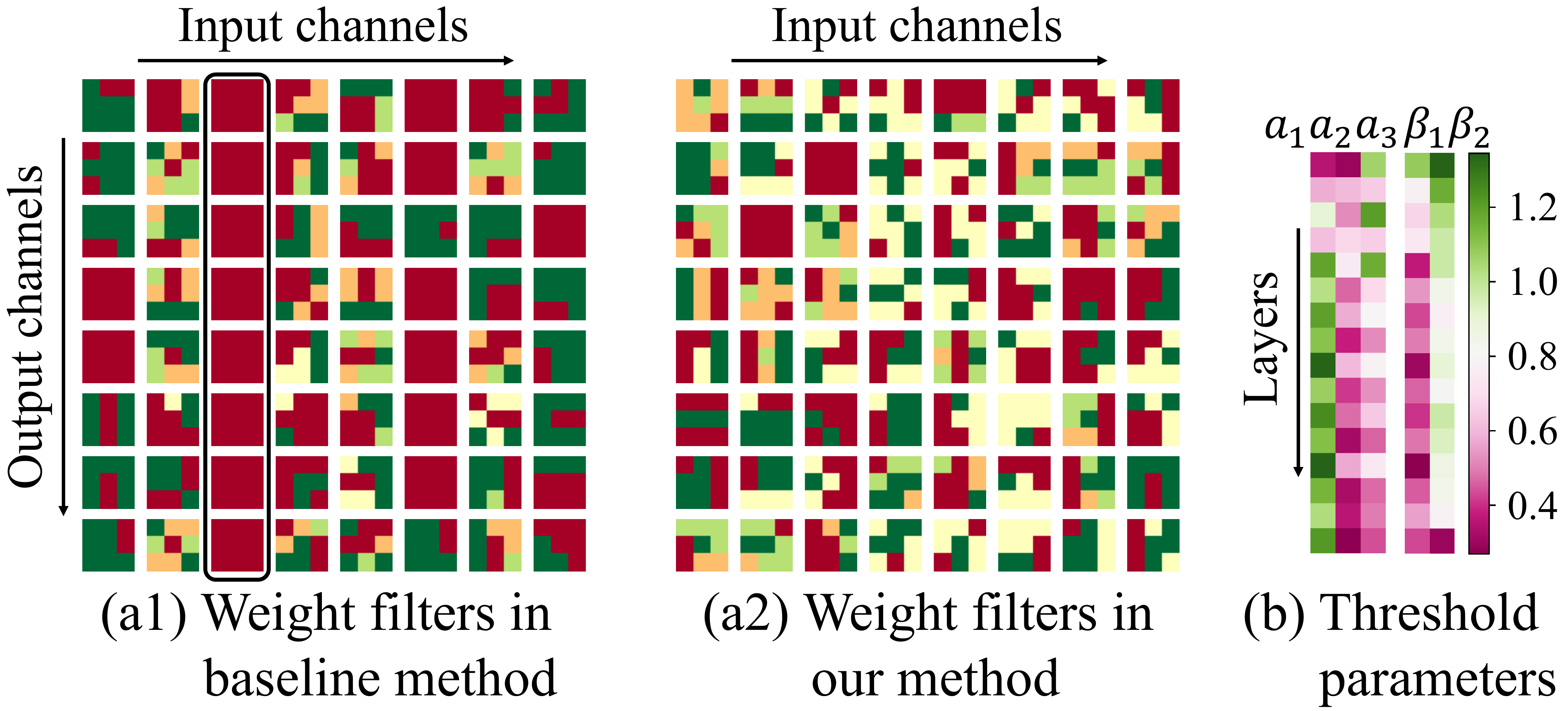}
\vspace{-0.5em}
\caption{Quantized weights and learnable thresholds visualization} 
\vspace{-0.5em}
\label{fig:weight_threshold_visualization}
\end{figure}

We visualize the optimized weights in the trained 2-bit ResNet-18. As shown in Fig.~\ref{fig:weight_threshold_visualization}, many $3\times3$ weight matrices learn the same value in the baseline method, which can hardly extract useful features. In contrast, this phenomenon is much rarer with the proposed weight regularization. Specifically, for 2-bit case, 4.37\% of quantized $3\times3$ weight matrices contain the same values in baseline method, while this number is reduced to 1.69\% in our method. 
Further, in Fig.~\ref{fig:weight_threshold_visualization} (b), the learned threshold parameters in the quantized ResNet-18 network have clear patterns. The parameters in first convolutional layers of the residual blocks (\textit{i.e.}, odd rows in Fig.~\ref{fig:weight_threshold_visualization} (b)) often have larger values in the threshold intervals $a$, and smaller values in the first scaling factors $\beta_1$. We deem it is because the real-valued activations in these layers are the summation of the residual connection and the previous layer output, thus are larger in magnitude, for which, larger $a$ and smaller $\beta_1$ are learned to better represent these activations in fixed bits.

\section{Results without Quantizing Downsampling Layers}
\label{sec:resnet_more}
In previous quantization works, there is a practice~\cite{liu2018bi, rastegari2016xnor} of keeping the down-sampling layers to be full-precision and quantizing the rest of the convolutional layers. We follow these studies and conduct experiments on ResNet. As shown in Table~\ref{table:resnet_more}, for lower bits, real-valued 1x1 downsampling layers can boost the accuracy for $\sim$0.3\%, while this effect becomes marginal for higher bits.

\setlength{\tabcolsep}{2pt}
\begin{table}[h]
\begin{center}
\caption{Accuracy comparison of quantizing the downsampling layers in ResNet. Both weights and activations are quantized to 2-bit 3-bit or 4-bit. * denotes keeping the weights and activations to be full-precision in 1$\times$1 downsampling layers and quantizing all the remaining convolutional and fully-connected layers except the first and the last one.}
\label{table:resnet_more}
\resizebox{0.46\textwidth}{!}{
\begin{tabular}{cccccccccccc}
\hline
\hline
\noalign{\smallskip}
\multirow{2}{*}{Network} & \multirow{2}{*}{Method} & \multicolumn{2}{c}{2-bit} & \multicolumn{2}{c}{3-bit} & \multicolumn{2}{c}{4-bit} \\
& &Top-1 & Top-5  & Top-1 & Top-5 & Top-1 & Top-5 \\
\noalign{\smallskip}
\hline
\multirowcell{2}{ResNet-18}
& N2UQ & 69.4 & 88.4 & 71.9 & 90.5  & 72.9 & 90.9\\
& \textbf{N2UQ*} & \textbf{69.7} & \textbf{88.9} & \textbf{72.1} & \textbf{90.5} & \textbf{73.1} & \textbf{91.2}\\
\hline
\multirowcell{2}{ResNet-34}
& N2UQ & 73.3 & 91.2 &75.2 & 92.3 & 76.0 & 92.8\\
& \textbf{N2UQ*} & \textbf{73.4} & \textbf{91.3} & \textbf{75.3} & \textbf{92.4} & \textbf{76.1} & \textbf{92.8}\\
\hline
\multirowcell{2}{ResNet-50}
& N2UQ & 75.8 & 92.3 & 77.5 & 93.6 & 78.0 & 93.9\\
& \textbf{N2UQ*} & \textbf{76.4} & \textbf{92.9} & \textbf{77.6} & \textbf{93.7} & \textbf{78.0} & \textbf{94.0}\\
\hline
\hline
\end{tabular}
}
\end{center}
\end{table}

\end{document}